\documentclass[journal]{IEEEtran}

\usepackage{amsmath,amsfonts,bm}

\def\eqref#1{equation~\ref{#1}}

\def\1{\bm{1}}

\DeclareMathAlphabet{\mathsfit}{\encodingdefault}{\sfdefault}{m}{sl}
\SetMathAlphabet{\mathsfit}{bold}{\encodingdefault}{\sfdefault}{bx}{n}

\usepackage{subfig}
\usepackage{epstopdf}
\usepackage{algorithm}
\usepackage{algorithmicx}
\usepackage{algpseudocode}
\usepackage{setspace}
\usepackage[T1]{fontenc}
\usepackage{color,soul}

\usepackage{hyperref}
\usepackage[final]{graphicx}
\usepackage{adjustbox}
\usepackage{float}
\usepackage{subfig}
\usepackage[algo2e,ruled,vlined,noend]{algorithm2e}
\usepackage{algpseudocode}
\usepackage{booktabs}
\usepackage{multirow}
\usepackage{url}

\title{Efficient Wrapper Feature Selection using Autoencoder and Model Based Elimination}

\author{Sharan~Ramjee, {\em Student Member, IEEE}, and Aly El~Gamal, {\em Senior Member, IEEE}
\thanks{S. Ramjee and A. El Gamal are with the Department of Electrical and Computer Engineering, Purdue University, West Lafayette, IN, USA. Email: \{sramjee, elgamala\}@purdue.edu.}
}

\begin{document}

\maketitle

\begin{abstract}
We propose a computationally efficient wrapper feature selection method - called Autoencoder and Model Based Elimination of features using Relevance and Redundancy scores (AMBER) - that uses a single ranker model along with autoencoders to perform greedy backward elimination of features.
The ranker model is used to prioritize the removal of features that are not critical to the classification task, while the autoencoders are used to prioritize the elimination of correlated features. We demonstrate the superior feature selection ability of AMBER on 4 well known datasets corresponding to different domain applications via comparing the accuracies with other computationally efficient state-of-the-art feature selection techniques. 
\end{abstract}
\section{Introduction}\label{intro}
Feature selection is a preprocessing technique that ranks the significance of features to eliminate features that are insignificant to the task at hand. As examined by \cite{yu2003feature}, it is a powerful tool to alleviate the curse of dimensionality, reduce training time and increase the accuracy of learning algorithms, as well as to improve data comprehensibility.
For classification problems, \cite{weston2001feature} divides feature selection problems into two types: $(a)$ given a fixed $k \ll d$, where $d$ is the total number of features, find the $k$ features that lead to the least classification error and $(b)$ given a maximum expected classification error, find the smallest possible $k$.
In this paper, we will be focusing on problems of type $(a)$. 
\cite{weston2001feature} formalizes this type of feature selection problems as follows. Given a function $y = f(x,\alpha)$, find a mapping of data $x \mapsto (x * \sigma)$, $\sigma \in \{0,1\}^d$, along with the parameters $\alpha$ for the function $f$ that lead to the minimization of
\begin{equation}
    \tau(\sigma,\alpha)=\int V(y,f((x*\sigma),\alpha))dP(x,y),
    \label{feat_eq}
\end{equation}
subject to $\Vert \sigma \Vert_0 = k$, where the distribution $P(x,y)$ - that determines how samples are generated - is unknown, and can be inferred only from the training set, $x * \sigma = (x_1\sigma_1,\dots,x_d\sigma_d)$ is an elementwise product, $V(\cdot,\cdot)$ is a loss function and $\Vert\cdot\Vert_0$ is the $L_0$-norm.

Feature selection algorithms are of $3$ types: Filter, Wrapper, and Embedded methods. 
Filters rely on intrinsic characteristics of data to measure feature importance while wrappers iteratively measure the learning performance of a classifier to rank feature importance.
\cite{Li_2017} asserts that although filters are more computationally efficient than wrappers, the features selected by filters are not as good.
Embedded methods use the structure of learning algorithms to embed feature selection into the underlying model to reconcile the efficiency advantage of filters with the learning algorithm interaction advantage of wrappers.
As examined by \cite{saeys2007review}, embedded methods are model dependent because they perform feature selection during the training of the learning algorithm.
This serves as a motivation for the use of wrapper methods that are not model dependent.
\cite{weston2001feature} define wrapper methods as an exploration of the feature space, where the saliency of subsets of features are ranked using the estimated accuracy of a learning algorithm. Hence, $\tau(\sigma, \alpha)$ in (\ref{feat_eq}) can be approximated by minimizing
\begin{equation}
    \tau_{wrap}(\sigma,\alpha)=\min_{\sigma} \tau_{alg}(\sigma),
    \label{wrapper_eq}
\end{equation}
subject to $\sigma \in \{0,1\}^d$, where $\tau_{alg}$ is a classifier having estimates of $\alpha$.
Wrapper methods can further be divided into three types: Exhaustive Search Wrappers, Random Search Wrappers, and Heuristic Search Wrappers. We will focus on Heuristic Search Wrappers that iteratively select or eliminate one feature at each iteration because unlike Exhaustive Search Wrappers, they are more computationally efficient and unlike Random Search Wrappers, they have deterministic guarantees on the set of selected salient features, as illustrated in \cite{hira2015review}.
\subsection{Motivation}\label{problem_intro}
\subsubsection{Relevance and Redundancy}\label{sec:relevance}
We hypothesize that the saliency of features is determined by two factors: Relevance and Redundancy. Irrelevant features are insignificant because their direct removal does not result in a drop in classification accuracy, while redundant features are insignificant because they are linearly or non-linearly dependent on other features and can be inferred - or approximated - from them as long as these other features are not removed.
As detailed by \cite{guyon2008feature}, one does not necessarily imply the other. 
Filter methods are better at identifying redundant features while wrapper methods are better at identifying irrelevant features, and this highlights the power of embedded methods as they utilize aspects of both in feature selection as mentioned in \cite{bolon2013review}.
Since most wrapper methods do not take advantage of filter method based identification of redundant features, there is a need to incorporate a filter based technique to identify redundant features into wrapper methods, which we address using autoencoders.

\subsubsection{Training the Classifier only once}
Wrapper methods often have a significantly high computational complexity because the classifier needs to be trained for every considered feature set at every iteration.
For greedy backward elimination wrappers, the removal of one out of $d$ features requires removing each feature separately and training the classifier with the remaining $d-1$ features and testing its performance on the validation set. 
The feature whose removal results in the highest classification accuracy is then removed. This is the procedure followed by most backward feature selection algorithms such as the Recursive Feature Elimination (RFE) method proposed by \cite{guyon2002gene}. 
For iterative greedy elimination of $k$ features from a set of $d$ features, the classifier has to be trained for $\sum_{i=1}^{k}(d-i+1)$ times, which poses a practical limitation when the number of features is large.
Also, the saliency of the features selected is governed by how good the classifier that ranks the features is, and as such, we need to use state-of-the-art classifiers for ranking the features (CNNs for image data, etc.).
These models are often complex and thus, consume a lot of training time which implies a trade-off between speed and the saliency of selected features.
We address this issue by training the feature ranker model only once.

\section{State of the art}\label{sec:sota}
We now describe top-notch efficient feature selection methods that we will be comparing our proposed method to. With the exception of FQI, the implementations of these methods can be found in the scikit-feature package
created by \cite{Li_2017}.

\textbf {Fisher Score} encourages selection of features that have similar values within the same class and distinct values across different classes. A precise definition is available in \cite{duda2012pattern}.

\textbf{Conditional Mutual Information Maximization (CMIM)} is a fast 
feature selection method proposed in \cite{vidal2003object} and \cite{fleuret2004fast} that iteratively selects features while maximizing the empirical Shannon mutual information function between the feature being selected and class labels, given already selected features. 
A precise definition is available in \cite{Li_2017}. 

\textbf{Efficient and Robust Feature Selection (RFS)} is an efficient feature selection method proposed by \cite{nie2010efficient} that exploits the
noise robustness property of the joint $\ell_{2,1}$-norm loss function, by applying the $\ell_{2,1}$-norm minimization on both the loss function and its associated regularization function. A precise definition is available in \cite{Li_2017}.
The value of the regularization coefficient for our experiments was chosen by performing RFS on a wide range of values and picking the value that led to the highest accuracy on the validation set.

\textbf{Feature Quality Index (FQI)} is a feature selection method proposed by \cite{de1997feature} that utilizes the output sensitivity of a learning model to changes in the input, to rank features. FQI serves as the main inspiration for our proposed method and as elaborated in \cite{verikas2002feature}, the FQI of feature $f_i$ is computed as\vspace{-0.05in}
\begin{equation}
\small
    FQI (f_i) = \sum_{j=1}^{n}\Vert o_j - o_j^i \Vert ^2,
\end{equation}
where $n$ is the total number of training examples, $o_j$ is the output of the model when the $j^{th}$ training example is the input, and $o_j^i$ is the output of the neural network when the $j^{th}$ training example, with the value of the $i^{th}$ feature set to $0$, is the input. 
\section{AMBER}\label{ddfs}
\subsection{Sensitivity of weights to features}\label{simulating}
Following a gradient-based optimization of a deep neural network, the weights connected to the neurons in the input layer that correspond to more salient features tend to have larger magnitudes and this has been extensively documented by \cite{bauer2000feature}, \cite{belue1995determining}, and \cite{priddy1993bayesian}. 
Similar to FQI, we measure the relevance of each feature by setting the input to the neuron corresponding to that feature to $0$.
This essentially means that the input neuron is dead because all the weights/synapses from that neuron to the next layer will not have an impact on the output of the neural network.
Since more salient features possess weights of higher magnitude, these weights influence the output to a greater extent and setting their values to $0$ in the input will result in a greater loss in the output layer.
This is the basis of the Weight Based Analysis feature selection methods outlined by \cite{lal2006embedded}.
We further note that we normalize the training set before training by setting the mean of each feature to $0$ and the variance to $1$, so that our \emph{simulation of feature removal} is effectively setting the feature to its mean value for all training examples.
To summarize, the pre-trained neural network ranker model prioritizes the removal of features that are irrelevant to the classification task by simulating the removal of each feature. Features whose removal results in a lower loss are less relevant. We will refer to the loss value of this ranker model as a feature's \textbf{Relevance Score}.
\subsection{Autoencoders Reveal Non-Linear Correlations}\label{autoencoder}
The weights connected to less salient features can possess high magnitudes, when these features are redundant in presence of other salient features as described in Sec. \ref{sec:relevance}.
Hence, we use a filter based technique that is independent of a learning algorithm to detect these redundant features.
We experimented with methods like PCA as detailed by \cite{witten2009penalized} and correlation coefficients as detailed by \cite{mitra2002unsupervised}
but these methods revealed only linear correlations in data. We hence introduced autoencoders into the proposed method because they reveal non-linear correlations as examined by \cite{han2018autoencoder}, \cite{balin2019concrete}, and \cite{Sakurada:2014:ADU:2689746.2689747}. To eliminate one feature from a set of $k$ features, we train the autoencoder with one dense hidden layer consisting of $k-1$ neurons using the normalized training set.
We note that this hidden layer can also be convolutional, LSTM, or of other types depending on the data we are dealing with.
To evaluate a feature, we set its corresponding values in the training set to $0$ and pass the set into the autoencoder. We then take the Mean Squared Error (MSE) between the output and the original input before the values corresponding to the evaluated feature were set to $0$, and perform this for each of the $k$ features separately.
Lower MSE values indicate higher likelihood for redundancy. We refer to this MSE as a feature's \textbf{Redundancy Score}.
\subsection{Using Transfer Learning to prevent retraining}
To eliminate $k$ out of $d$ features, we pick a state-of-the-art neural network model for the dataset and train it on the training set using part of it as the validation set. We call this model the \textbf{Ranker Model} (RM) as it allows us to rank the saliency of the features. Next, we set the input for each of the $d$ features in all the examples of the training set to $0$ one at a time in a round-robin fashion to obtain a list of $d$ Relevance Scores. 
Additionally, we train the autoencoder and pass the same modified training sets through the autoencoder to obtain $d$ Redundancy Scores.
We then divide the Relevance and Redundancy Scores by their corresponding ranges so that they both contribute equally to the final decision and add them to obtain a \textbf{Saliency Score}.
The feature with the lowest Saliency Score is eliminated from the training set. In the context of the RM, elimination means that the feature is permanently set to $0$ for all the examples in the training set, as we reuse the same RM in further iterations.
In the context of the autoencoder, elimination means that that feature is permanently removed from the training set for all the examples. This entire process, without retraining the RM, is done iteratively $k$ times.
The pseudocode for AMBER is described in Algorirthm \ref{algo:pseudocode}. We note that a feature selection algorithm can be implemented either to remove a specified number of features or to stop when the accuracy of the learning algorithm ceases to increase. We chose the former method to take into account scenarios when this accuracy decreases before it increases later on in the feature selection process. Further, it is straightforward to implement a variant of AMBER according to the latter objective. 

\begin{algorithm}[H]
\SetAlgoNoLine
\DontPrintSemicolon
\textbf{Inputs}: k: Number of features to be eliminated; trainSet: Training Dataset;\\
\textbf{Outputs}: featList: List of $k$ eliminated features
\\[0.05in]
{\bf function} AMBER(k, trainSet)\\
    Train state of the art RM using trainSet\\
    Initialize featList to empty list\\
    \For{$i = 1$ to $k$}
        
       Set rmSet as trainSet with all features in featList set to $0$\\
        Set autoTrainSet as trainSet where all features in featList are removed\\
        Train autoencoder with one hidden layer containing $d-i$ units using autoTrainSet\\
         \For{$j$ in $d-i+1$ features not in featList}
             
             Record loss of RM when rmSet is evaluated after setting feature $j$ to $0$\\
            Set cTrainSet as autoTrainset where feature $j$ is set to $0$\\
            Record MSE of autoTrainSet and output of autoencoder when cTrainSet is evaluated\\[0.05in]
            \textbf{End for}\\
         Normalize RM losses and MSEs and add corresponding values\\
         Sort and add lowest scoring feature to featList\\
         \textbf{End for}\\
    {\bf return} (featList)
\caption{AMBER Algorithm for Feature Selection}
\label{algo:pseudocode}
\end{algorithm}

\section{Results}\label{others}
\subsection{Experimental Setup}
We used $3$ Nvidia Tesla P100 GPUs, each with $16$ GB of memory, and Keras with a TensorFlow backend. With the exception of the RadioML2016.10b dataset for which we used all $3$ GPUs, we only used $1$ GPU for training.
The experiments were performed $3$ times and the average accuracies were plotted at each feature count in Fig. \ref{fig:comparison_plots}.
\footnote{The source code for AMBER, links to the datasets considered, and the error bars for the comparison plots are available at \url{https://github.com/alyelgamal/AMBER}}.
\subsection{Datasets and Classifiers}\label{sec:dataset}
Each dataset corresponds to a different application domain to demonstrate the versatility of AMBER.
The final models that are trained with the set of selected features are common across all the feature selection methods that are compared and are trained until early stopping is achieved with a patience value of $5$ to ensure that the comparisons are fair.
For all the datasets, the softmax activation function is applied to the output layer with the cross-entropy loss function, and ReLU is applied to hidden layers, unless explicitly stated otherwise. All the autoencoders used have a dense hidden layer. We note that we obtained slightly better results with a convolutional and LSTM hidden layer for the MNIST and RadioML2016.10b datasets, respectively.
The test split used for the Reuters and the Wisconsin Breast Cancer datasets is $0.2$ while the test split used for the RadioML2016.10b dataset is $0.5$.
Some of the plots in Fig. \ref{fig:comparison_plots} were jagged when feature counts in decrements of $1$ were plotted and thus, we plotted them in larger feature count decrements.
Finally, to demonstrate that the final model does not necessarily have to be the same as the RM used by AMBER, we used different models as the final model and the RM for the MNIST and RadioML2016.10b datasets.

{\bf MNIST} is a handwritten recognition dataset created by \cite{lecun1998mnist} consisting of $60000$ $28$x$28$ grayscale images with $10$ classes, each belonging to one of the $10$ digits, along with a test set that contains $10000$ images of the same dimensions. The total number of features is $784$.
The Ranker Model is a CNN consisting of $2$ convolutional layers, a max pooling layer, and $2$ dense layers, in that order. The convolutional layers have $32$ and $64$ filters, in order of depth, with kernel sizes of ($3$x$3$) for both layers. The max pooling layer has a ($2$x$2$) pool and the dense layers have $128$ and $10$ (output layer) neurons. 
The final model used is an MLP model consisting of $3$ fully connected layers with $512$, $512$, and $10$ (output layer) neurons. Each of the hidden layers is followed by a 0.2 rate dropout layer.

{\bf Reuters} is a Keras built-in text dataset that consists of $11228$ newswires from Reuters with $46$ classes, each representing a different topic. Each wire is encoded as a sequence of word indices, where the index corresponds to a word's frequency in the dataset. For our demonstration, the $1000$ most frequent words are used.
The Ranker Model and the final model are the same MLP model consisting of $2$ fully connected layers with $512$ and $46$ (output layer) neurons.  

{\bf Wisconsin Breast Cancer} is a biological dataset created by \cite{street1993nuclear} that consists of features that represent characteristics of cell nuclei that have been measured from an image of Fine Needle Aspirates (FNAs) of breast mass. The dataset consists of $569$ examples that belong to $2$ classes: malignant and benign. The total number of features is $30$.
The Ranker Model and the final model are the same MLP model consisting of $4$ fully connected layers with $16$, $8$, $6$, and $1$ (output layer) neurons, in order of depth. Sigmoid activation is used for the output layer.


{\bf RadioML2016.10b} is a dataset of received wireless signal samples used by \cite{conv} that consists of $1200000$ $128$-sample complex time-domain vectors with $10$ classes, representing different modulation types.
It consists of $20$ Signal to Noise Ratios (SNR) ranging from -$20$ dB to $18$ dB in increments of $2$ dB; we only choose the results of the $18$ dB data for better illustration.
Each of the $128$ samples consists of a real part and a complex part and thus, the input dimensions are $2$x$128$, and the total number of features is $256$. This dataset has the unique property that only pairs of features (belonging to the same sample) can be eliminated. AMBER, like FQI, is powerful in such situations as the pairs of features can be set to $0$ to evaluate their collective rank.
The other feature selection methods fail in this case because they account for feature interactions within the pairs of features as well, which is one reason for why AMBER outperforms them as it does not.
For the other methods, to eliminate pairs of features belonging to the same sample, we simply added the scores belonging to the two features to obtain a single score for each sample.
The Ranker Model used here is a CLDNN while the final model used is a ResNet, both of which are described in \cite{ramjee2019fast}.

\subsection{Classification Accuracies}
\begin{figure}[H]
    \captionsetup[subfigure]{labelformat=empty}
    \centering
    \subfloat[(a)]{{\includegraphics[width=\columnwidth]{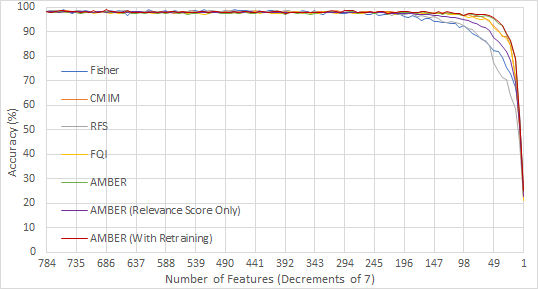}}}%
    \qquad \qquad 
    \subfloat[(b)]{{\includegraphics[width=\columnwidth]{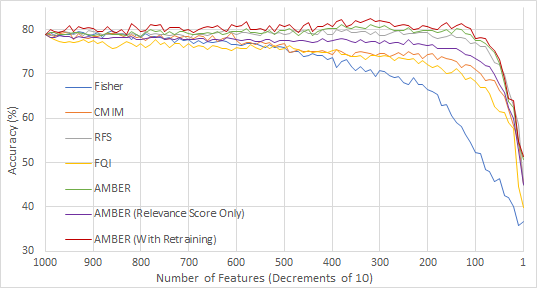}}}
	\qquad \qquad
	\subfloat[(c)]{{\includegraphics[width=\columnwidth]{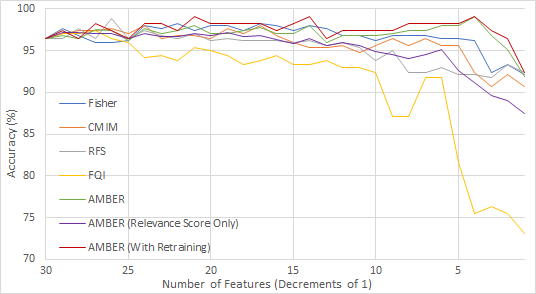}}}%
    \qquad \qquad 
    \subfloat[(d)]{{\includegraphics[width=\columnwidth]{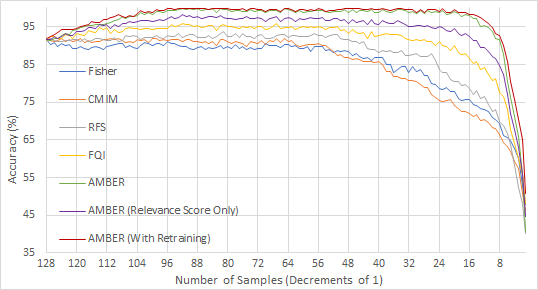}}}
    \caption{Accuracy vs Feature Count plots for the final models trained with the selected features for the (a) MNIST, (b) Reuters, (c) Wisconsin Breast Cancer, and (d) RadioML2016.10b datasets.}
	\label{fig:comparison_plots}
\end{figure}

The final models' classification accuracy plots using the selected features can be observed in Fig. \ref{fig:comparison_plots}.
We observe the impressive performance delivered by AMBER that generally outperforms that of all $4$ considered methods, particularly when the number of selected features becomes very low (about $99\%$ average accuracy with $4$ out of $30$ features for the Cancer dataset and about $95\%$ average accuracy with $16$ out of $128$ samples for the RadioML dataset).
The comparisons of the accuracies of the final models using the top $10\%$ of features 
are given in Table \ref{tab:comparisons}. The results in the last two rows refer to using a version of AMBER without the Autoencoder's redundancy score, and another version of AMBER where the ranker model is retrained in every iteration, respectively. Note from the depicted results (purple curves in the figure) how solely relying on the RM significantly reduces AMBER's performance, which validates our intuition about the benefit of using the Autoencoder to capture correlations to reduce the generalization error. Further, we observe how negligible gains are achieved when retraining the RM in every iteration, that comes at a significant computational cost, as demonstrated in Table~\ref{tab:time_comparisons}, which validates our intuition about simulating the removal of features without retraining for computational efficiency while maintaining good performance.   
\begingroup
\setlength{\tabcolsep}{5pt}
\begin{table}[H]
    \caption{Accuracy Comparisons}
    \label{tab:comparisons}
    \centering
    \small
    \begin{tabular}{lllll}
        {\bf Method} &\multicolumn{4}{c}{{\bf Avg. accuracy with top ${\bf 10\%}$ features (\%)}}\\
        \cmidrule(r){1-5}
        &{\bf MNIST} &{\bf Reuters} &{\bf Cancer} &{\bf RadioML}\\
        \cmidrule(r){2-5}
        \centering
        Fisher &88.37 &51.21 &92.40 &73.42\\
        CMIM &96.38 &71.04 &90.64 &70.22\\
        RFS  &89.46 &77.11 &91.81 &75.85\\
        FQI &95.49 &68.20 &76.32 &83.57\\
        \textbf{AMBER} &\bf{97.21} &\bf{77.55} &\bf{96.78} &\bf{95.15}\\
        - Relevance &93.29 &73.45 &89.65 &89.54\\
        - Retraining &97.21 &78.11 &97.37 &97.49\\
    \end{tabular}
\end{table}
\endgroup

\begin{table}[H]
    \caption{Time needed to rank all features in Seconds.}
    \label{tab:time_comparisons}
    \centering
    \small
    \begin{tabular}{lllll}
        {\bf Method}&{\bf MNIST} &{\bf Reuters} &{\bf Cancer} &{\bf RadioML}\\
        \centering
        AMBER &10552.24 &21710.78 &40.04 &26417.53\\
        - Retraining &24202.66 &29005.08 &739.01 &42533.27\\
    \end{tabular}
\end{table}

\section{Discussion}
\subsection{Feature Selection leads to higher accuracies}
In some cases, like in the cases of the Wisconsin Breast Cancer and the RadioML datasets, we observed that with AMBER, the accuracy of the final model trained with the selected subset of features is higher than the model trained with all the features.
We believe that in most cases where this happens, it is because the model was overfitting the data as training examples belonging to different classes were compactly packed in the feature space consisting of all the features.
However, once they were projected onto the feature space consisting of the selected subset of features, the same data could be better distinguished as a decision boundary, with the same complexity as before, divided these training examples better.
\vspace{-4mm}
\begin{figure}[H]
    \captionsetup[subfigure]{labelformat=empty}
    \centering
    \subfloat[PCA before AMBER]{{\includegraphics[width=0.4\columnwidth]{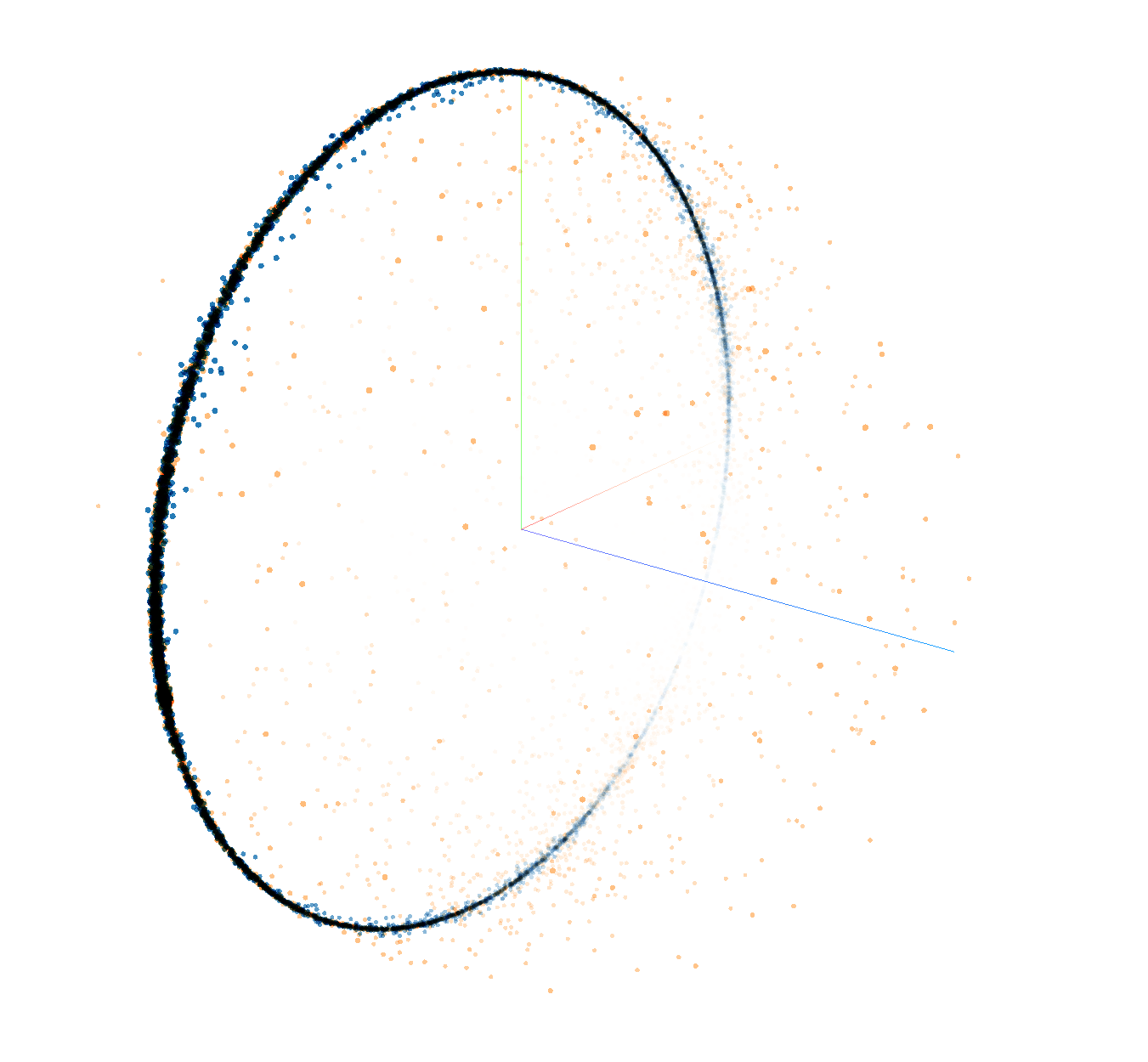}}}%
    \qquad  
    \subfloat[PCA after AMBER]{{\includegraphics[width=0.4\columnwidth]{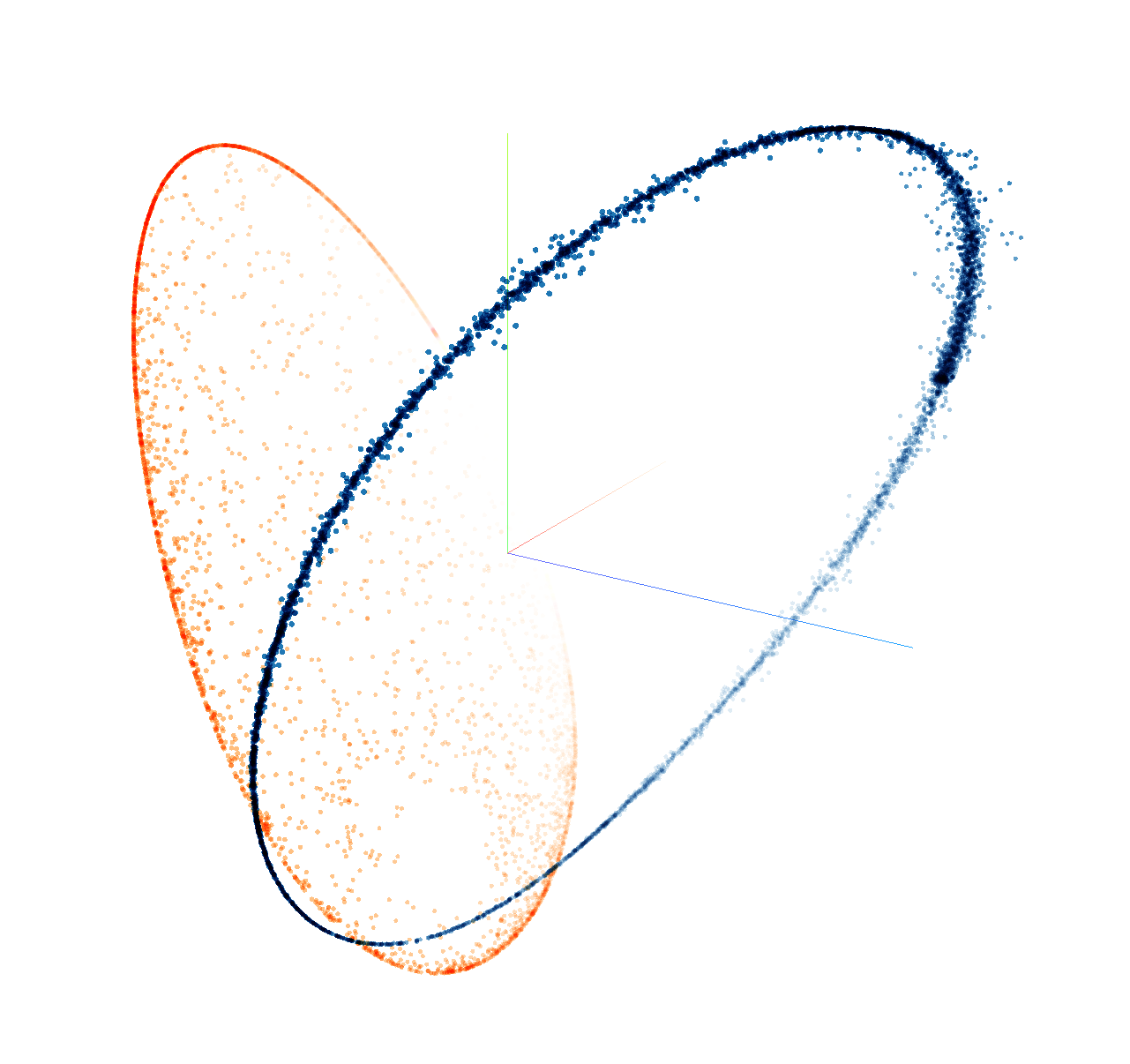}}}
    \caption{PCA 3-D visualizations of the training dataset for the AM-DSB (blue) and WBFM (orange) classes before and after AMBER where $64$ samples ($128$ features) were selected.}
	\label{fig:pca_viz}
\end{figure}
For instance, in the case of the RadioML2016.10b dataset, the accuracy with all the $128$ samples ($256$ features) was about $93\%$, where the main source of error was the AM-DSM and WBFM classes that were often misclassified. After AMBER is used to reduce the number of samples to $64$, the accuracy increased to about $99\%$. To illustrate this, we used PCA to reduce the dimensions of the training set that belong to these two classes to $3$-D and plotted the training set before and after AMBER in Figure \ref{fig:pca_viz}.

\subsection{Overfitting the RM facilitates better Feature Selection}
In Section \ref{ddfs}, we elaborated on how more salient features possess higher magnitudes of weights in the input layer than features that are less salient, which is the property of neural networks that serves as the basis for AMBER.
The performance of AMBER heavily depends on the performance of the RM that ranks the features. In some cases, however, even the state of the art models do not have high classification accuracies. In such cases, we can obtain better feature selection results with AMBER by overfitting the RM on the training set. \vspace{-5mm}
  \begin{figure}[H]
    \centering
    \subfloat[\label{subfig-1:dummy}]{%
          \includegraphics[width=\columnwidth]{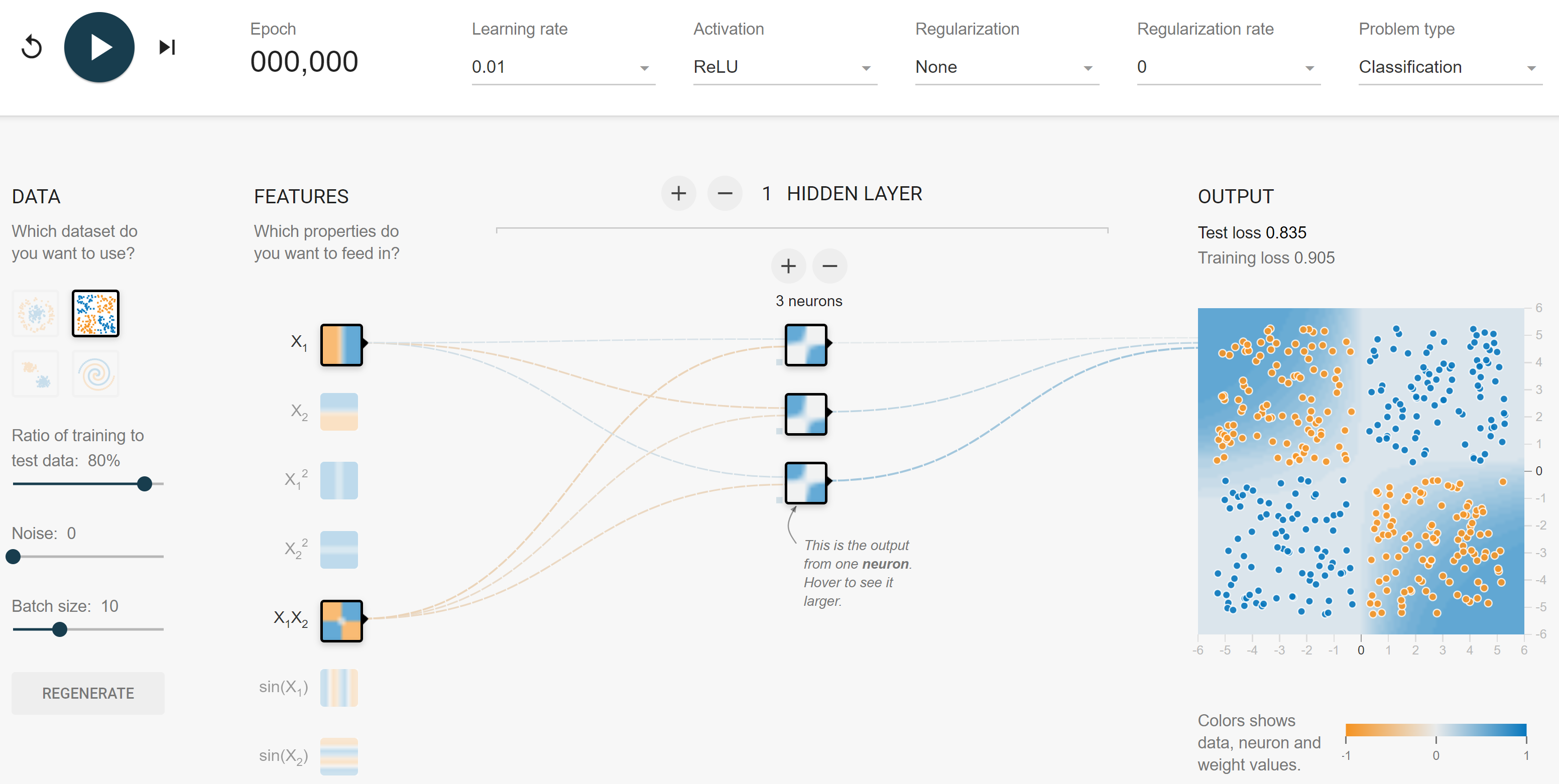}}

    \subfloat[\label{subfig-2:dummy}]{%
          \includegraphics[width=.5\columnwidth]{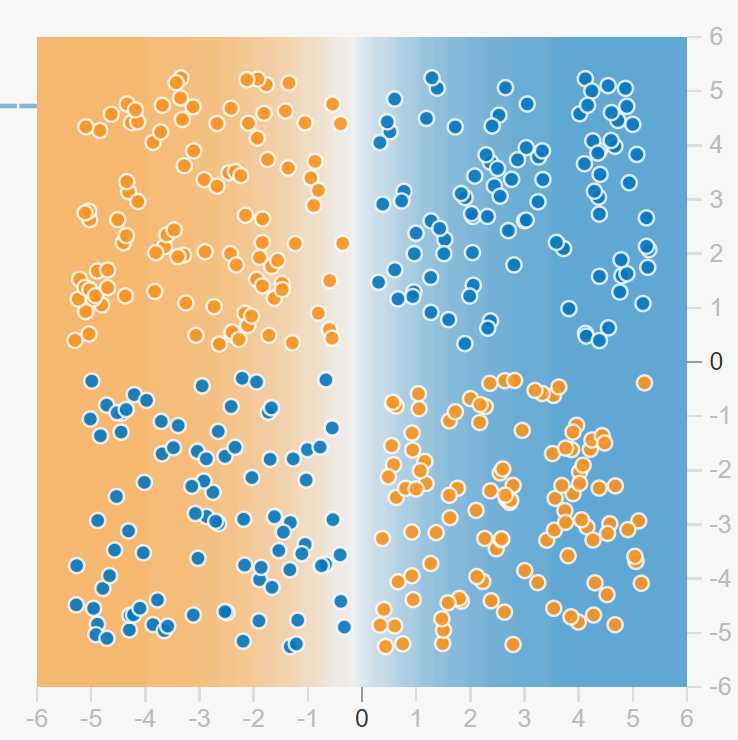}} 
    \subfloat[\label{subfig-3:dummy}]{%
          \includegraphics[width=.5\columnwidth]{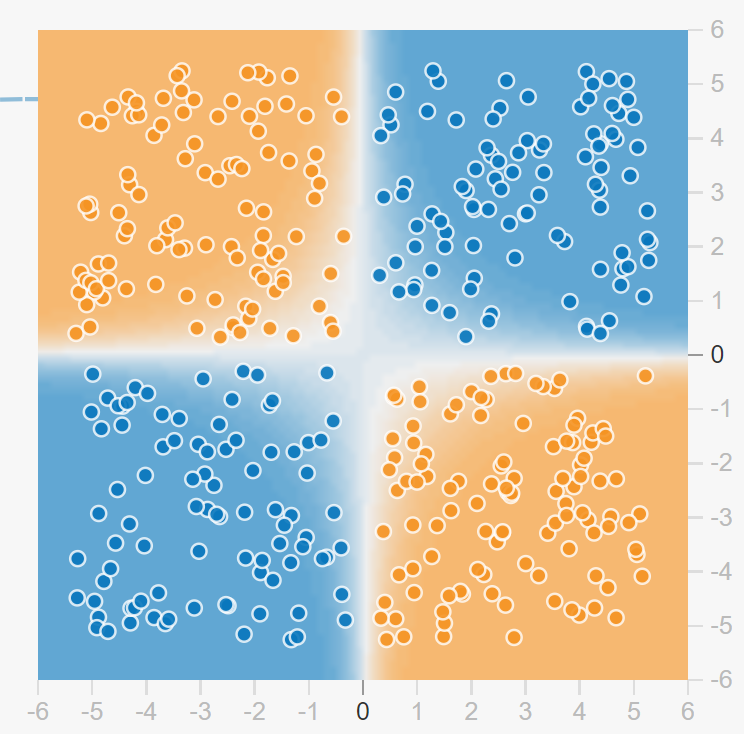}}
    
    \caption{(a): Toy example from TensorFlow Playground; (b) Feature $1$ and (c) Feature $2$.}\label{fig:toy_screenshot}
  \end{figure}
  
  \begin{figure}
    \captionsetup[subfigure]{labelformat=empty}
    \centering
    \subfloat[(a)]{{\includegraphics[width=0.7\columnwidth]{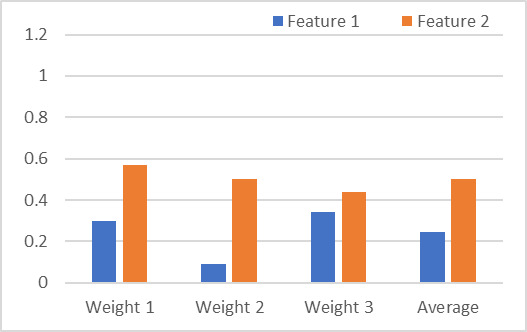}}}
    
    \subfloat[(b)]{{\includegraphics[width=0.7\columnwidth]{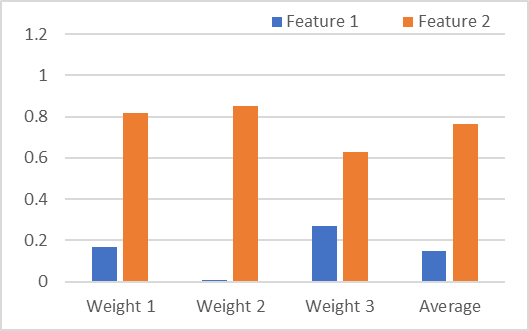}}}
    
    \subfloat[(c)]{{\includegraphics[width=0.7\columnwidth]{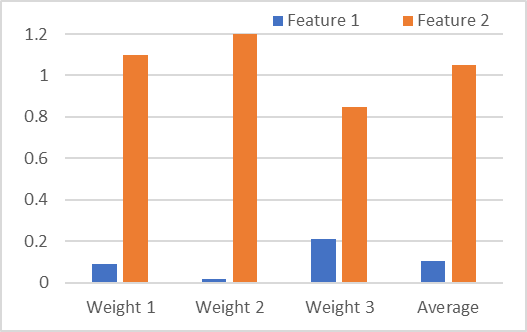}}}
    \caption{Input layer weight magnitudes after training for (a) 10, (b) 100, and (c) 1000 epochs.\vspace{-5mm}}
	\label{fig:toy_overfitting}
\end{figure}

We demonstrate the insight behind this using the toy example illustrated in Fig. \ref{fig:toy_screenshot} that portrays the architecture used for the RM along with the corresponding hyperparameters.
Here, feature $1$ is $x_1$ and feature $2$ is $x_1x_2$. Feature $2$ is more salient than feature $1$ as it enables a decision boundary that allows for better classification of the data (shown in (c)), while feature $1$ cannot.
Each of these features have $3$ weights in the input layer and as expected, the weights connected to feature $2$ manifest into weights of higher magnitude than those belonging to feature $1$ as shown in Fig. \ref{fig:toy_overfitting}.
As the number of training epochs increases, the difference in the average magnitudes of the weights increases. 
This would imply a greater difference between the Relevance scores of these two features. 
Thus, we can overfit the RM on the training set by training for a large number of epochs without regularization to make the magnitudes of the weights connected to more salient features in the input layer even larger than those connected to less salient features.

\section{Concluding Remarks}
AMBER presents a valuable balance in the trade-off between computational efficiency in feature selection, in which filter-based methods excel at, and performance (i.e. classification accuracy), in which traditional wrapper methods excel at. It is inspired by FQI with two major differences: $1$- Instead of making the final selection of the desired feature set based on simulating the model's performance with elimination of only a single feature, the final model's performance in AMBER is simulated with candidate combinations of selected features, $2$- The autoencoder is used to capture redundant features; a property that is missing in FQI as well as most wrapper feature selection methods. However, we found AMBER to require slightly larger computational time than the considered 4 state of the art methods, and we also found it to require far less time than state of the art wrapper feature selection methods, as it does not require retraining the RM in each iteration.
It is also worth mentioning that the final values that the weights connected to the input features manifest after training are dependent on the initialization of these weights. We believe - and plan to investigate in future work - that following carefully designed initialization schemes could allow us to create a larger difference between the magnitudes of the weights of more salient and less salient features.

\bibliographystyle{IEEEtran}
\bibliography{refs}

\end{document}